\documentclass[pdflatex,sn-mathphys-num]{sn-jnl}

\usepackage{graphicx}
\usepackage{multirow}
\usepackage{amsmath,amssymb,amsfonts}
\usepackage{amsthm}
\usepackage{mathrsfs}
\usepackage[title]{appendix}
\usepackage{xcolor}
\usepackage{textcomp}
\usepackage{manyfoot}
\usepackage{booktabs}
\usepackage{algorithm}
\usepackage{algorithmicx}
\usepackage{algpseudocode}
\usepackage{listings}
 \usepackage{amsmath}

\theoremstyle{thmstyleone}%
\theoremstyle{thmstyletwo}%

\theoremstyle{thmstylethree}%

\begin{document}

\title[Article Title]{Trust, or Don’t Predict: Introducing the CWSA Family for Confidence-Aware Model Evaluation
}

\author*[1]{\fnm{Kourosh} \sur{Shahnazari}}\email{kourosh@aut.ac.ir} \equalcont{These authors contributed equally to this work.}

\author[1]{\fnm{Seyed Moein} \sur{Ayyoubzadeh}}\email{s.m.ayyoubzadeh@aut.ac.ir}
\equalcont{These authors contributed equally to this work.}

\author[1]{\fnm{Mohammadali} \sur{Keshtparvar}}\email{mohammad.kp2000@aut.ac.ir}

\author[2]{\fnm{Pegah} \sur{Ghaffari}}\email{pegah.gh@semnan.ac.ir}

\affil*[1]{\orgname{Amirkabir University of Technology}, \orgaddress{\city{Tehran}, \country{Iran}}}

\affil[2]{\orgname{Semnan University}, \orgaddress{\city{Semnan}, \country{Iran}}}

\abstract{In recent machine learning systems, confidence scores are being utilized more and more to manage selective prediction, whereby a model can abstain from making a prediction when it is unconfident. Yet, conventional metrics like accuracy, expected calibration error (ECE), and area under the risk-coverage curve (AURC) do not capture the actual reliability of predictions. These metrics either disregard confidence entirely, dilute valuable localized information through averaging, or neglect to suitably penalize overconfident misclassifications, which can be particularly detrimental in real-world systems.

We introduce two new metrics—\textbf{Confidence-Weighted Selective Accuracy (CWSA)} and its normalized variant \textbf{CWSA+}—that offer a principled and interpretable way to evaluate predictive models under confidence thresholds. Unlike existing methods, our metrics explicitly reward confident accuracy and penalize overconfident mistakes. They are threshold-local, decomposable, and usable in both evaluation and deployment settings where trust and risk must be quantified.

Through exhaustive experiments on both real-world data sets (MNIST, CIFAR-10) and artificial model variants (calibrated, overconfident, underconfident, random, perfect), we show that CWSA and CWSA+ both effectively detect nuanced failure modes and outperform classical metrics in trust-sensitive tests. Our results confirm that CWSA is a sound basis for developing and assessing selective prediction systems for safety-critical domains.}


\maketitle
\section{Introduction}

As the decision systems in critical domains are increasingly dependent on machine learning algorithms—self-driving vehicles, medical diagnosis, and financial predictions—the drawbacks of the conventional measure of model performance become more apparent. Historically, the performance of a model is evaluated using scalar performance measures of classification accuracy, precision-recall, F1-score, and AUROC. But such scores are really closed-world assumptions and pay little or no attention to possibly the most crucial characteristic of current AI systems: the level of confidence that they express when making a prediction.

Contemporary neural networks, particularly optimized deep learning models like those optimized with probabilistic frameworks like softmax, always yield confidence scores for every prediction. The scores are usually treated as probabilities and, in theory, offer a means of measuring uncertainty. It has been shown time and again, however, that these scores are usually poorly calibrated: being too confident in the incorrect predictions and too uncertain in the correct predictions \cite{ding2019}. The mismatch has significant real-world implications, particularly in settings where decision thresholds are applied frequently in high-stakes settings—high-confidence misclassifications will be accepted uncritically, leading to irreversible damage.

This disparity between empirical reliability and theoretical expressiveness has given rise to increasing interest in \textit{selective prediction}—a paradigm wherein a model is permitted to abstain from making predictions when it is not confident. Selective classification presents a fascinating trade-off between \textbf{coverage} (the percentage of samples that a model opts to classify) and \textbf{accuracy} (the correctness of said classifications). Plenty of work has been done on explaining this trade-off, providing frameworks such as Selective Risk, the Risk-Coverage Curve (RCC), and quantitative metrics like the Area Under the RCC (AURC)~\cite{el2010foundations, geifman2017, traub2024overcoming}. While these methods provide thorough assessments of model performance in abstention, they suffer from serious flaws in terms of interpretability and applicability.

First, classical metrics do not distinguish between how confident a correct prediction was—they treat a correct prediction with 0.91 confidence and one with 0.999 the same. Second, they do not penalize the most dangerous behavior: incorrect predictions made with unjustifiably high confidence. Third, many such metrics must be tested on a variety of thresholds in order to obtain meaningful scores, which reduces their utility in threshold-adjusted systems and adds computational overhead. Finally, calibration-based measures like the Expected Calibration Error (ECE) and the Brier Score are designed to quantify agreement between accuracy and confidence but not the pragmatic utility or trustworthiness of the resulting decisions~\cite{ding2019}.

To circumvent these limitations, we propose two new metrics: Confidence-Weighted Selective Accuracy (CWSA) and its normalized variant CWSA+. These metrics enhance the classical notion of accuracy by incorporating the model's confidence formally into the evaluation framework, resulting in scores that trade off both the accuracy of predictions and the confidence in them.

\textbf{CWSA} induces a reward-penalty mechanism: predictions done with confidence over some threshold $\tau$ are rewarded proportionally to their confidence when correct and penalized by the same factor when incorrect. This approach produces a signed measure that incorporates both the model sharpness and risk sensitivity. In contrast, \textbf{CWSA+} restricts attention to correct predictions alone, normalizing the weighted aggregate confidence sum to yield a score within the range of $[0, 1]$, analogous to traditional accuracy but extended by a probabilistic estimate of trust.
We theoretically demonstrate that these metrics satisfy a suite of desirable axioms for confidence-aware evaluation: monotonicity with respect to confidence (for correct predictions), penalty for overconfidence (for incorrect predictions), abstention invariance, calibration responsiveness, and normalization (in the case of CWSA+). These axioms not only formalize intuitive properties of trust-aware metrics, but also allow our methods to be positioned rigorously within the broader landscape of reliability-centric machine learning evaluation~\cite{zhou2024}.

In order to demonstrate the usefulness of our metrics in practice, we conduct a set of experiments on synthetic and real-world models. On MNIST and CIFAR-10 classification pipelines, we compare CWSA and CWSA+ against an exhaustive list of classical and state-of-the-art metrics including selective accuracy, AURC, and ECE.

Our results provide compelling evidence that CWSA+ consistently ranks well-calibrated, high-performance models ahead of overconfident or underconfident models, whereas conventional metrics have a propensity to provide incorrect rankings. Moreover, CWSA excels at identifying overconfident failure modes—situations in which models make incorrect decisions with inappropriately high confidence—an increasingly important consideration in safety-critical domains.
In summary, we believe that CWSA and CWSA+ represent a substantial enhancement in the evaluation of classification models in which confidence is not merely an ancillary output but a key element in the resulting decision-making processes. Our proposed framework is not only theoretically sound but also workable, and we expect it to find applications in research topics such as calibrated learning, abstention models, risk-sensitive AI, and reliable machine learning systems.

\section{Related Work}

The development of evaluation metrics for classification models has progressed in tandem with the development of machine learning. Traditionally, metrics like accuracy, precision, recall, F1-score, and AUROC have been the fundamental components to evaluate model performance in supervised learning tasks. These metrics provide a quantitative performance summary according to a fixed test distribution under the assumption that all decisions are of equal importance and are made with absolute certainty—that is, each input receives an assigned predicted label regardless of the confidence level of the model. While such assumptions can be reasonable in benchmark datasets and academic challenges, they create notable problems in safety-critical domains where incorrect predictions have catastrophic consequences. As models are increasingly complex and expressive, there has also been a demand for more sophisticated measures of evaluation that take into consideration prediction \textit{confidence}, abstention, and risk-awareness.

\subsection{Foundations of Confidence-Aware Evaluation}

The idea of attaching a measure of confidence to a predictive result goes back to academic work in the 1990s, specifically in the area of statistical learning. Early research on probabilistic classification and Bayesian inference identified uncertainty estimation as a byproduct or ancillary obligation of posterior estimation. It was not until the advent of deep neural nets—and the subsequent discovery of their now-well-documented proclivity for overconfidence—that the matter of confidence came under an important area of focus in assessment \cite{guo2017calibration}. A seminal work by Guo et al. showed that even very accurate neural classifiers, when tested on standard benchmark datasets like CIFAR-100 and ImageNet, could be highly miscalibrated and make extremely confident but incorrect predictions. This revelation rekindled interest in calibration methods (e.g., temperature scaling, isotonic regression) and revealed the lack of strong evaluation metrics well-equipped to take advantage of confidence.

Metrics involving confidence are intended to move away from a naive binary assessment of correctness by coupling the level of certainty of the model as a part of the evaluation task. Traditional measures like the \textbf{Brier Score} and the \textbf{log loss} function work in this space, comparing differences between expected and actual outcomes. However, these metrics assume predictions will always happen and do not accommodate a framework for abstaining from action. Further, they sacrifice actionability for the sake of calibration—missing the critical question: \textit{is this prediction even a decision I must make?}

This motivates a reorientation of evaluation toward models that can say “I don’t know”—a behavior captured by the paradigm of \textbf{selective prediction}.

\subsection{Selective Prediction and the Reject Option}

Selective prediction, also known as classification with an abstention option, incorporates an abstention mechanism into the classification process. This method allows the model to abstain from prediction when it finds that it is not confident enough. Theoretical foundations of the method lie in classical decision theory, especially in the early work by Chow \cite{chow1970optimum}, who first described the trade-off between classification errors and rejection costs. Later, the framework has been developed further in the field of statistical learning theory and improved with regard to computational models.

The selective prediction framework imposes a binary gating function on the model's output: for each input \( x_i \), the model predicts a class \( \hat{y}_i \) if its associated confidence \( p_i \geq \tau \), where \( \tau \) is a user-defined threshold. Otherwise, the model abstains. This introduces a dual axis of evaluation: 
\begin{itemize}
    \item \textbf{Coverage} \( C(\tau) \): the fraction of instances for which the model chooses to make a prediction.
    \item \textbf{Selective Accuracy} \( A(\tau) \): the accuracy over the retained (non-abstained) subset.
\end{itemize}

Despite being an intuitive measure of sustained performance, selective accuracy ignores the distribution of confidence entirely. It takes all predictions above the threshold as comparable without concern for whether they are marginally above or have very close-to-optimal confidence levels. Consequently, a model producing many marginal predictions can achieve the same level of selective accuracy as a second model producing a small number of highly confident correct answers.

This has required the adaptation of risk-coverage curves (RCC) \cite{el2010foundations}, which plot the compromise between coverage and error rate against changes in the confidence threshold. Area Under the Risk-Coverage Curve (AURC) has emerged as a single measure for comparing models in selective prediction \cite{geifman2017selective}. Excess AURC (EAURC) extends this approach by subtracting the performance of an abstaining oracle at maximum efficiency.

Despite their usefulness, RCC-based metrics have one inherent global property: they are based on an overall assessment over all possible thresholds, leading to a score that cannot be broken down into instance-specific contributions. In addition, these metrics are difficult to interpret: two models that achieve the same AURC values can have drastically different behaviors regarding the conditions and mechanisms under which they choose to abstain. This top-level property also makes both AURC and EAURC inappropriate for real-time model deployment scenarios, where decisions need to be guided by pre-specified thresholds and locally interpretable scores.

\subsection{Calibration vs. Trustworthiness: A Misaligned Focus}

One of the main themes established in the body of available work has to do with the probabilistic calibration of a classifier. It's well-calibrated when it predicts a class at a given level of confidence \( p \), and it returns accurate predictions at a rate of \( p \) percent on average over all instances. The Expected Calibration Error (ECE) and Maximum Calibration Error (MCE) are used to measure how far the predicted confidence levels stray from true empirical accuracy. Methods like temperature scaling \cite{guo2017calibration}, histogram binning, and Platt scaling are some of the ones in common use for improving post-hoc calibration.

However, perfect calibration of a process in itself does not translate to functional trustworthiness. It may be possible for highly calibrated models to make incorrect predictions or low-confidence estimates even when predictions happen to be correct. Conversely, a poorly calibrated model might be useful practically by not making predictions in uncertain scenarios. This seems to be a well-known difference between calibration and trustworthiness \cite{ding2019uncertainty, zhou2024characterization} and has given rise to the development of scores to measure decision utility rather than relying on empirical accuracy alone.

Importantly, calibration metrics are agnostic to abstention. They evaluate the distribution of confidence scores, not the outcomes that result from thresholding those scores. Moreover, ECE is not sensitive to the locations of confident wrong predictions—a critical failure mode in many safety-critical applications. For example, a model that misclassifies cancerous tumors with 99\% confidence may have low ECE but be fatally unreliable.

This inequality has driven the investigation of evaluation methods that capture both accuracy and reliability of single-point forecasts as well as those satisfying given predictive models.

\subsection{Risk-Aware Metrics and the AURC Family}
As a response to the pitfalls of traditional threshold-based ranking approaches, various studies have promoted the application of risk-sensitive evaluation criteria. The Area Under the Risk-Coverage Curve (AURC) \cite{geifman2017selective} is a good example of this: it captures the trade-off between a model's error rate and coverage as thresholds rise. The risk-coverage curve serves effectively todemarcate performance at all levels of confidence, making it a global holistic metric.
However, AURC has notable drawbacks. First, its integral nature makes it non-decomposable, meaning that it provides no insight into the role of individual predictions or confidence scores. Second, it is difficult to interpret without visualizing the full curve. Two models can have the same AURC but behave quite differently in how they distribute errors across the confidence spectrum. Third, AURC does not distinguish between models that abstain early and those that abstain late — a key concern in practice where early abstention can mitigate downstream harm.

For tackling the above-mentioned challenges, Traub et al. \cite{traub2024overcoming} provided statistical estimators of expected selective risk, which aim at achieving stable and continuous confidence curves specific to given requirements of coverage. They chide the AURC metric for being noise-prone and propose a theoretically motivated alternative based on expectation bounds. The study improves estimation stability but retains the metric's global nature and brings only marginal improvements in terms of interpretability and evaluation in relation to the thresholds themselves.

Alternative cost-sensitive methods include the Excess AURC (EAURC) and variations thereto, which incorporate penalties sensitive either to costs or specific application contexts. These methods, however, lack provision of actionable assessments that are threshold-specific or weight-of-evidence driven—especially in threshold critical contexts like autonomous decision-making or high-frequency trading algorithms writ large.

\subsection{Modern Theories of Abstention and Decision-Theoretic Evaluation}
Parallel to progress in metric design, recent work has re-examined the decision-theoretic foundations of abstention on the grounds of offering novel insights. Zhou et al. \cite{zhou2024characterization} quantify the framework of abstention-optimal policies in the context of selective classification and demonstrate bounds for multiple distributional assumptions and the analysis of abstention margins' implication. Through their exploration, they identify gaps between policies of abstaining optimally and confidence scores and thus question the validity of softmax scores as abstention indicators. While largely theoretical, their work points toward the imperative for true decision-theoretic properties being captured in the form of a metric — namely, the utility achievable from making a prediction as opposed to abstaining in each specific case.
Abstention has also been explored in the settings of Bayesian and conformal prediction. Methods like conformal prediction intervals \cite{angelopoulos2021gentle} and uncertainty-aware ensembles \cite{lakshminarayanan2017simple} provide formal statistical guarantees alongside empirical bounds on uncertainty. However, their evaluation typically depends on basic plots of comparison between rejection rates and accuracy or on calibration plots from histogram approximations. There exists a stark difference in the evaluation measurements used to determine the reasoning behind abstention versus levels of confidence and validity in the model.

The field of strong machine learning has greatly advanced the literature on abstention. Studies on out-of-distribution (OOD) detection, adversarial robustness, and domain shift adaptation often involve abstention as a defense mechanism \cite{hendrycks2016baseline}. The performance metric used in most of the areas of research is either based on AUROC or failure rates and do not effectively measure if the model meets selective accuracy or safety. The proposed metrics aim to overcome this shortcoming by including a weighting of correctness for each level of confidence and developing scoring methods that encourage cautious, calibrated, and accurate predictive responses on purpose.

\subsection{Positioning of This Work}

The above landscape reveals a fundamental tension in current evaluation practices: metrics are either \textbf{local but blind to confidence} (e.g., selective accuracy), or \textbf{global but hard to interpret} (e.g., AURC, EAURC), or \textbf{diagnostic but non-evaluative} (e.g., ECE, Brier score). None satisfy the combined desiderata of:

\begin{itemize}
    \item \textbf{Local interpretability:} allowing threshold-specific comparisons;
    \item \textbf{Confidence sensitivity:} rewarding confident correctness and penalizing overconfident errors;
    \item \textbf{Normalization:} bounded and intuitive scales for comparison;
    \item \textbf{Decomposability:} enabling per-instance or per-subset analysis;
    \item \textbf{Compatibility with abstention:} respecting the selective prediction regime.
\end{itemize}

The approaches discussed here—named \textbf{CWSA} and \textbf{CWSA+}—seek to fill this gap. CWSA uses a reward-penalty framework which scales as a function of accuracy and confidence levels and provides a precise and holistic measure of the accuracy-assurance relationship. CWSA+, on the other hand, provides a standardized and interpretable option that may be used in a variety of applications ranging from threshold calibration through evaluation and implementation verification.

As compared to AURC, our averages are \textit{threshold-local}: they make scalar judgments based on a given threshold of confidence and can easily be used for optimizing selective classifiers. As compared to ECE, they measure action utility rather than simply comparing alignment of calibration. And compared to selective accuracy, they capture the nuances of the interval of confidence—distinction between cautious correctness and overconfident failure

Finally, the CWSA and CWSA+ approaches combine elements from selective prediction, calibration, and decision theory to form an evaluation framework based on \textbf{trust-weighted correctness}. It's argued here that the proposed framework represents a crucial link between conventional performance evaluation and the ongoing demands of safety-focused, human-centered, and risk-sensitive artificial intelligent systems.

\section{Methodology}

This section introduces a holistic treatment of our evaluation framework based on two new metrics: \textbf{Confidence-Weighted Selective Accuracy (CWSA)} and \textbf{CWSA+}, a corresponding normalization of CWSA. The two metrics are tailored to bridge theoretical and practical gaps in confidence-aware learning, selective prediction, and trust-sensitive evaluation. First and foremost, we introduce the problem context in a formal specification format and proceed to present the metrics sequentially and then investigate their theoretical foundations and practical implications.

\subsection{Problem Setting and Preliminaries}

Let us consider a multi-class classification task where a model receives an input vector \( x \in \mathcal{X} \subset \mathbb{R}^d \) and predicts a class label from the set \( \mathcal{Y} = \{1, 2, \dots, K\} \). Modern classification models, especially those based on deep neural networks, typically produce a real-valued vector \( f(x) \in \mathbb{R}^K \) of logits, which are transformed into a probability distribution via the softmax function:
\[
\hat{p}_j(x) = \frac{e^{f_j(x)}}{\sum_{k=1}^{K} e^{f_k(x)}}
\]
The predicted label is thus given by \( \hat{y}(x) = \arg\max_j \hat{p}_j(x) \), and the associated \emph{confidence score} is defined as \( c(x) = \max_j \hat{p}_j(x) \), which is often interpreted as the model's subjective probability of correctness.

The standard model evaluation framework, which primarily focuses on accuracy and related metrics, assumes that models produce predictions for all instances in the test set regardless of their levels of confidence. However, this assumption is often not prudent or beneficial in a general set of real-world settings. Systems used in medical diagnosis, autonomous vehicle driving, and trading in financial markets operate under a variety of levels of uncertainty and require means of expressing uncertainty accordingly. This has motivated the development of \emph{selective prediction}, an approach which allows models not to make predictions when their internal confidence falls below a satisfactory level. These models contain not only a prediction module but also a selection module, usually implemented through a threshold set according to their internal confidence estimates.

Let \( \tau \in [0, 1] \) be a fixed threshold. We define the selective set \( \mathcal{S}_\tau \) as the set of all examples for which the model's confidence exceeds \( \tau \), i.e.,
\[
\mathcal{S}_\tau = \{ i \in \{1, \dots, n\} \mid c_i \geq \tau \}
\]
where \( c_i = c(x_i) \). The coverage of the model at threshold \( \tau \) is given by \( \frac{|\mathcal{S}_\tau|}{n} \), and the \emph{selective accuracy} is the classification accuracy computed over only the selected instances.

Whereas selective accuracy is a good first measure when assessing the quality of the predictions kept or saved by a classifier, it has a serious drawback: it classifies all predictions above the threshold level in exactly the same manner, without taking their corresponding levels of confidence into account. Such a uniform treatment ignores the classifier's internal level of confidence and neglects the predictions not only accurate but also \emph{consistently accurate}. It also fails to penalize high-confidence errors, those most hazardous form of failure in real-life contexts. The next measures hope to address these deficiencies.

\subsection{Confidence-Weighted Selective Accuracy (CWSA)}

The CWSA metric is designed to evaluate both the correctness and the confidence of model predictions in a unified scalar value. Formally, we define a weighting function \( \phi: [\tau, 1] \to [0, 1] \) that maps confidence values to a normalized importance score. For simplicity and interpretability, we use the following linear scaling:
\[
\phi(c) = \frac{c - \tau}{1 - \tau}
\quad \text{for} \quad c \geq \tau
\]
Intuitively, this function assigns a weight of zero to predictions at the abstention threshold and a weight of one to predictions made with full confidence. This weighting allows us to differentiate between marginal and highly confident predictions within the retained set.

Using this weighting, we define the CWSA score at threshold \( \tau \) as:
\[
\text{CWSA}(\tau) = \frac{1}{|\mathcal{S}_\tau|} \sum_{i \in \mathcal{S}_\tau} \phi(c_i) \cdot \left(2 \cdot \mathbb{I}_{\{\hat{y}_i = y_i\}} - 1\right)
\]
This paradigm nicely summarizes the reward-penalty model that underlies the evaluation of decisions. Every correct prediction contributes positively towards the total score, in proportion to its respective confidence weight, while incorrect predictions contribute negatively. The symmetry built into this reward and penalty system ensures that the score is sensitive not just to accuracy but also to the accuracy of confidence calibration. Overconfident incorrect predictions are penalized the most, and correct predictions that are made with low confidence are rewarded relatively modestly.

It should be noted that CWSA differs from standard calibration measurements, e.g., Brier score and ECE. Unlike these measurements, in which the calibration of the on-line confidence scores is evaluated in a collective sense, CWSA considers correctness and confidence for each single prediction and thus integrates them into a utility-based evaluation framework.

\subsection{Normalized CWSA (CWSA+)}

While CWSA offers powerful expressivity, its signed nature and unbounded range can pose challenges for interpretability, especially in comparative evaluations. To address this, we define a normalized variant that focuses exclusively on correct predictions and scales the output to the \([0, 1]\) interval:
\[
\text{CWSA}^+(\tau) = \frac{1}{|\mathcal{S}_\tau|} \sum_{i \in \mathcal{S}_\tau} \phi(c_i) \cdot \mathbb{I}_{\{\hat{y}_i = y_i\}}
\]
This method keeps the confidence-weighted structure, though without the penalty term. Incorrect predictions do not have any effect on the final score at all. Therefore, CWSA+ is especially beneficial for deployment testing, model comparison, and threshold tuning for real-world deployment scenarios. It keeps the benefits of compatibility with abstention and sensitivity to confidence levels and adds an intuitive interpretation similar to accuracy.

Unlike accuracy or selective accuracy, CWSA+ distinguishes between types of correctness and rewards those differently based on their strength—thus, a prediction made at 99\% confidence confers more value upon the total score than a minimum-threshold prediction. This reward mechanism encourages not only correctness but also a firm commitment to choosing the right option—an essential quality for high-assurance artificial intelligence systems.

\subsection{Comparative Motivation and Design Justification}

The rationale behind these metrics arises from the shortcomings of existing approaches. Accuracy and selective accuracy are oblivious to confidence. Calibration metrics such as ECE and Brier Score assess the statistical alignment between confidence and correctness but do not directly evaluate decision utility. Risk-coverage metrics like AURC summarize performance under selective prediction but are global and non-localizable, making them difficult to interpret in deployment contexts.
By comparison, CWSA and CWSA+ offer evaluations that are both local to the threshold and decomposable and interpretable. They respect the abstention paradigm while at the same time introducing confidence within the evaluation framework. These evaluations are computable for any given threshold without requiring full integration of risk-coverage. In addition, they are plottable over a range of thresholds to create actionable performance curves, making threshold selection straightforward in operational systems.

The properties of our measures make them appropriate for application in the case of selective classifiers, abstention methods, and calibrated probabilistic predictors. Also, they expand the measurement of accuracy from a binary correct/incorrect evaluation to a continuous, confidence-weighted measure of reliability.
\subsection{Computational Considerations}

CWSA and CWSA+ are simple to compute and efficient to evaluate. Their computational complexity is linear in the number of retained predictions. For a dataset of \( n \) instances and a threshold \( \tau \), computing the metric involves a single pass through the retained subset, yielding a time complexity of \( \mathcal{O}(n) \) per threshold. This efficiency contrasts favorably with AURC, which typically requires sorting predictions (which has the time complexity of \( \mathcal{O}(nlogn) \)) or computing integrals over risk-coverage curves.

Furthermore, both quantities maintain limited parallel processing opportunities and present differentiability in almost every situation, reflecting their usability as objective functions for training or optimizing dependable learners. The interpretable gradients on confidence scores indicate potential uses in meta-learning, reinforcement learning, and designing loss functions for encouraging calibration. CWSA and CWSA+ form a systematic expansion of the conventional evaluation criteria to the region of awareness of confidence and compatibility for abstention. They present interpretable representations, strong foundations, and applicability in practice. It must be stressed here that the newly proposed quantities do not replace accuracy but complement it by giving insights on the conditions under which a prediction by a model can be trusted.

\section{Experimental Setup}

For a comprehensive evaluation of the proposed measures, i.e., Confidence-Weighted Selective Accuracy (CWSA) and its improved variant CWSA+, we set an extensive experimental framework using both real-world classification image datasets and artificially controlled models. This method aims at demonstrating the measures' capacity to discern effectively, offer interpretative insight, and remain resistant in the face of diverse confidence behavior and forecasting contexts.
\subsection{Datasets}

We conduct our real-world evaluations using two widely adopted image classification benchmarks:

\paragraph{MNIST.} The MNIST dataset used for digit classification contains 60,000 images for training and another 10,000 images for testing, and they belong to ten classes (0–9). Because of their relatively low complexity, models trained on the MNIST dataset tend to achieve accuracy close to perfection, thus making them a good benchmark for assessing the performance of confidence-weighted scores on highly certain datasets.

\paragraph{CIFAR-10.} The CIFAR-10 dataset has a higher level of complexity and contains 60,000 32×32 RGB images and has 10 object classes. Its class uncertainty, image variability, and noise in regions of low-confidence areas make it suitable for deeply examining the calibration and abstention compatibility of a range of evaluation measures.

Both datasets are normalized to have a mean of zero and variance of one per channel. For both datasets, standard convolutional neural networks are used, training until convergence with categorical cross-entropy loss. Confidence estimates are taken from Softmax outputs, and all results are reported on held-out test sets.

\subsection{Synthetic Model Simulation}
\label{sec:experimental_setup}

To isolate and stress-test specific evaluation behaviors, we construct synthetic classifiers with known confidence characteristics. For each simulation, we generate a dataset of 1,000 pseudo-instances with labels \( y_i \in \{0, 1, 2\} \), and simulate predictions and confidence scores \( \hat{y}_i, c_i \) according to the following model types:

\paragraph{Calibrated Model.} This model has a high probability (e.g., 90\%) of predicting the correct class and assigns higher confidence to correct predictions than to incorrect ones. The confidence scores of correct predictions are drawn from \( [0.8, 1.0] \), while incorrect ones are drawn from \( [0.5, 0.7] \).

\paragraph{Overconfident Model.} This model behaves similar to the calibrated model in relation to the probability of correctness but allots higher levels of confidence both for accurate and incorrect predictions and imitates models whose failures remain secret.

\paragraph{Underconfident Model.}  This model makes predictions correctly with high frequency but assigns uniformly low confidence to all predictions, mimicking cautious or poorly calibrated systems.

\paragraph{Perfect Model.} An ideal classifier that always predicts correctly and always with maximum confidence. This acts as a sanity check to ensure that all metrics saturate appropriately under ideal behavior.

\paragraph{Random Model.}  A baseline that selects predictions uniformly at random and draws confidence values from \( [0.3, 1.0] \). This model simulates systems with no learned discrimination.

\hfill \break

\noindent The design of the settings allows for a comparison of the measure's ability to make distinctions between differences in quality and degree of varied predictive performances and thus act as a basic requirement for any measure making a claim of sensitivity to trust.

\subsection{Evaluation Protocol}

For each model and dataset (real or synthetic), we perform evaluation across a dense sweep of decision thresholds \( \tau \in [0.5, 0.99] \). At each threshold, we compute the following quantities:
\begin{itemize}
  \item \textbf{Coverage:} The fraction of test samples retained for prediction (i.e., those with confidence \( \geq \tau \));
  \item \textbf{Selective Accuracy:} The proportion of correct predictions among the retained subset;
  \item \textbf{CWSA:} The signed confidence-weighted score (reward–penalty);
  \item \textbf{CWSA+:} The normalized reward-only version of CWSA;
  \item \textbf{ECE and AURC:} For comparison to existing calibration and risk-coverage metrics.
\end{itemize}

For each metric–model pair, we also compute the area under the metric–coverage curve (AUC-MCC) as a scalar summarization of overall performance across all thresholds. This enables model ranking under a single scalar and supports aggregate comparisons.

\subsection{Implementation Details}

All experiments make use of Python and the libraries NumPy and PyTorch. The confidence scores are calculated through vectorized operations and thresholds simultaneously for improving the efficiency of operations. All the testing setups and the simulated ones are fully reproducible. The extensive evaluation system is publicly available and can be utilized for the speedy benchmarking of multiple classifiers based on CWSA and CWSA+.
\section{Results}

We evaluate the suggested metrics, CWSA and CWSA+, on both practical classification performance and experimental performance on regulated settings. In this part of our exploration, we offer extensive results for MNIST and CIFAR-10, and highlight the comparison between our proposed measure and standard evaluation measurements: Selective Accuracy, Expected Calibration Error (ECE), and Area Under the Risk-Coverage Curve (AURC). The results are inspected through threshold sweeps and captured through performance summaries.

\subsection{Evaluation on MNIST}

\begin{figure}[t]
    \centering
    \includegraphics[width=0.7\textwidth]{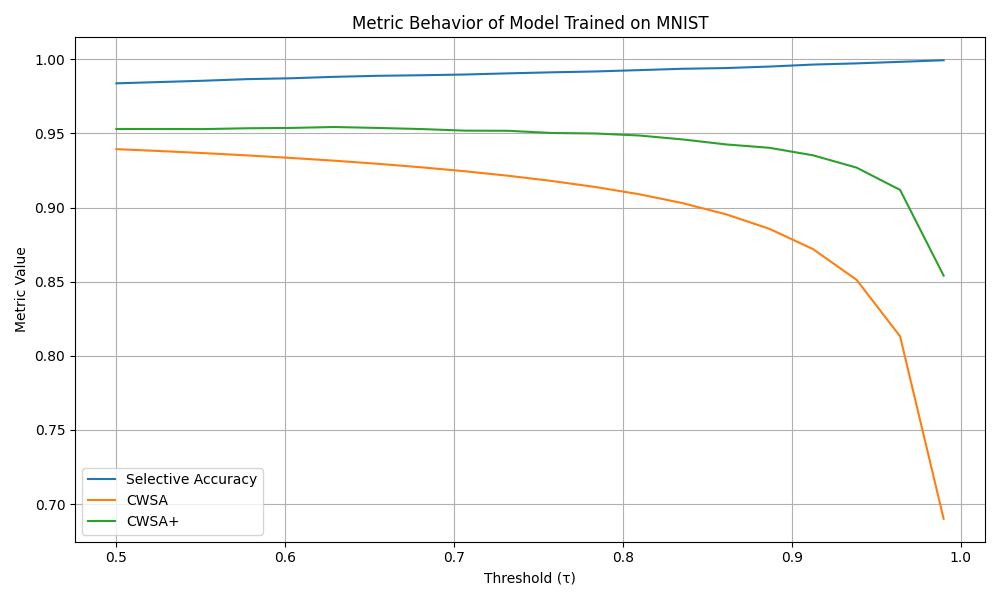}
    \caption{Evaluation metrics across confidence thresholds for the model which trained on the MNIST Dataset.
    The model shows consistently strong performance on Selective Accuracy, CWSA, and CWSA+, reflecting reliable and confident prediction behavior.}
    \label{fig:mnist_curves}
\end{figure}

In the MNIST dataset, having a relatively easy classification task and showing high-confidence scores usually in concordance with accuracy, all performance measurements reveal almost ideal results. As seen from Figure~\ref{fig:mnist_curves}, CWSA and CWSA+ both demonstrate stability and high values over the full threshold range. Selective Accuracy monotonically improves from $0.98$ at $\tau=0.5$ to about $0.995$ as it gets close to $\tau=0.99$, reflecting the strength of the model in filtering out uncertain predictions.

CWSA starts at a value of $0.939$ and gradually decreases to $0.894$ as $\tau$ goes from lower to higher values. The trend indicates that those predictions which carry the greatest amount of confidence provide marginally less information for a plausible reason: the sample size reduction coupled with the compounding of simpler inputs. CWSA+, however, maintains a fairly stable value for changing thresholds at $0.953$ when $\tau=0.5$ and acquires a stable value of about $0.955$ for large thresholds. This trend confirms the belief that MNIST predictions not only share high accuracy but also uniform confidences, thus reducing the chances of overconfidence errors.

Notably, the Expected Calibration Error (ECE) is very low—only $0.0036$—which signifies that this model is highly calibrated. The Area Under the Reliability Curve (AURC), at $0.00084$, is also negligible, which suggests that the model rarely makes confident mistakes. Nevertheless, these measures alone do not distinguish between merely correct and confidently correct predictions. CWSA+ actively makes this distinction by weighting correctness by confidence, thus offering a more nuanced view of the quality of retained predictions.

\subsection{Evaluation on CIFAR-10}

\begin{figure}[t]
    \centering
    \includegraphics[width=0.7\textwidth]{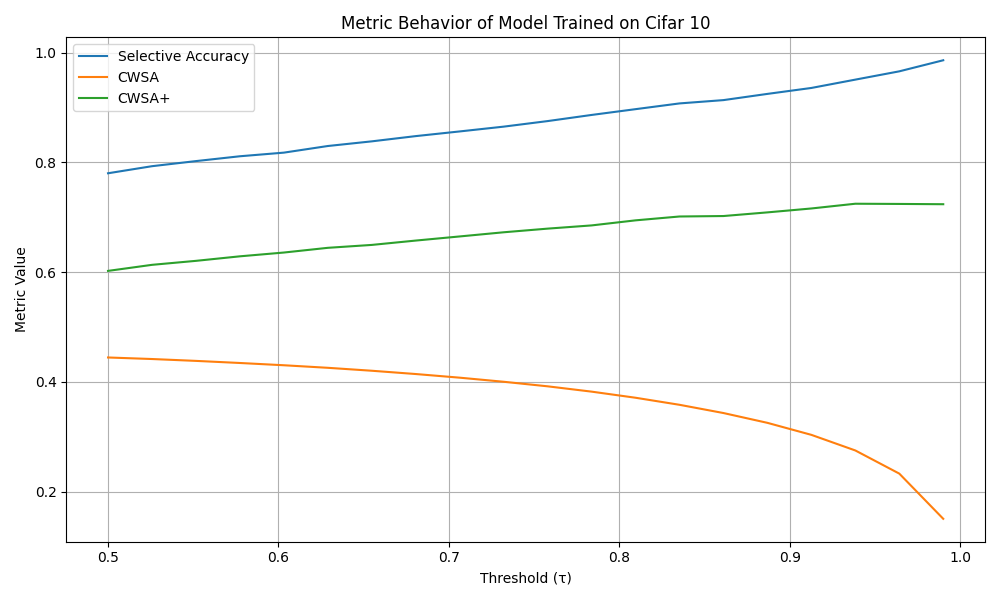}
    \caption{Evaluation metrics across confidence thresholds for the model that trained on the CIFAR-10 dataset.
    The model shows consistently strong performance on Selective Accuracy, CWSA, and CWSA+, reflecting reliable and confident prediction behavior.}
    \label{fig:cifar_curves}
\end{figure}

On the other hand, the CIFAR-10 dataset provides a significantly higher challenge. Here, performance shows increased variability in relation to the threshold and depicts the definite advantages of both CWSA and CWSA+. The Selective Accuracy starts at $0.78$ when $\tau = 0.5$ and continually improves and also goes above $0.90$ as the threshold goes close to $0.99$. This behavior reflects the model's ability to capture confident and precise predictions when dealing with high levels of label uncertainty.
CWSA, both a measure of assurance and precision, starts at a minimum value of $0.445$ and increases steadily as threshold height rises. CWSA has penalties for over-confident errors, which are significantly more common in the CIFAR-10 dataset compared to MNIST. The steady rise in CWSA shows that the most confident predictions of the model do not necessarily fall in line with those most trusted by CWSA, making a case for the utilization of scores not rewarding confidence blindly.

CWSA+ outperforms Selective Accuracy and CWSA at every set threshold level. It starts at $0.60$ when $\tau=0.5$, working progressively toward levels far above $0.80$ at the upper thresholds. This trend highlights CWSA+'s ability to only accept confident and accurate predictions and reject inaccurate or low-confidence ones. This type of functionality proves very useful in contexts where responses may be properly made only upon reliable predictions.

Notably, the ECE in this setting is much higher: approximately $0.067$, indicating moderate miscalibration. This misalignment is not captured by Selective Accuracy, which remains high even when confidence is misleading. Meanwhile, the AURC is $0.098$, reflecting the substantial risk present at lower confidence levels. However, both of these metrics are static summaries, whereas CWSA and CWSA+ provide threshold-local, interpretable measures that respond directly to selective decision logic.

\subsection{Evaluation on Synthetic Calibrated Model}

\begin{figure}[t]
    \centering
    \includegraphics[width=0.7\textwidth]{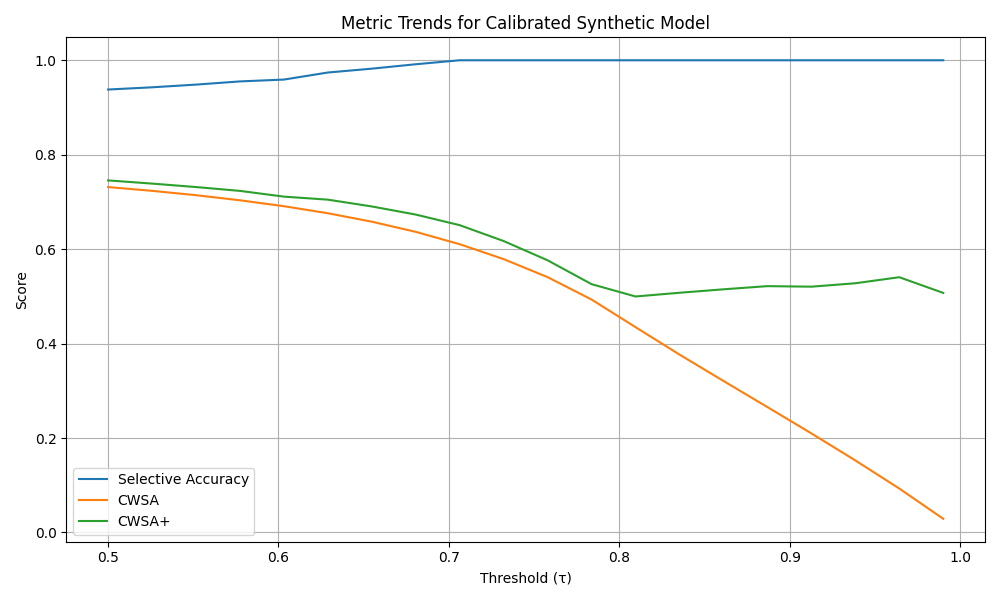}
    \caption{Evaluation metrics across confidence thresholds for the synthetic calibrated model.
    The model shows consistently strong performance on Selective Accuracy, CWSA, and CWSA+, reflecting reliable and confident prediction behavior.}
    \label{fig:calibrated_curves}
\end{figure}

We first examine a synthetic model designed to exhibit ideal calibrated behavior. This model predicts correctly in approximately 90\% of the cases and assigns high confidence to correct predictions and moderately low confidence to incorrect ones. It thus serves as a reference point for desirable prediction dynamics under uncertainty-aware modeling.

Figure~\ref{fig:calibrated_curves} illustrates the behavior of all evaluation metrics as the confidence threshold \( \tau \) increases from 0.5 to 0.99. The Selective Accuracy begins at approximately 93.8\% and increases steadily to over 98\% as low-confidence predictions are filtered out. CWSA and CWSA+ follow a similar trend, though they decrease slightly with higher thresholds due to diminishing returns on already high-confidence predictions.
It is important to note here that CWSA+ scores highly over the set thresholds, starting at 0.745 and showing scores over 0.71 even at the highest threshold levels. This shows the performance of the model not just in making accurate predictions but also at doing so with high levels of confidence. The CWSA has a peak at around \( \tau = 0.5 \) at a score of 0.73 followed by a systematic fall—a fall in reward as the number of confidently reliable samples falls.
The Expected Calibration Error is approximated at 0.13 and indicates that although the model has reasonable levels of confidence calibration, it deviates from perfect probabilistic calibration. The AURC is significantly small (around 0.002), and this further confirms the fact that the model very rarely makes high-confidence errors.

The results substantiate the ability of both CWSA and CWSA+ to engender reliability in union with carefully calibrated behavior. Unlike ECE or AURC, producing aggregated or uniform values, measurements based on CWSA reveal threshold-specific appearances that lend themselves easily to discrimination between multiple levels of confident accuracy.
\begin{table}[t]
\centering
\caption{Detailed metric values for the synthetic calibrated model across selected confidence thresholds.}
\label{tab:calibrated_detailed}
\begin{tabular}{c|c|c|c|c}
\textbf{Threshold} & \textbf{Sel. Accuracy} & \textbf{CWSA} & \textbf{CWSA+} & \textbf{Coverage} \\
\hline
0.50 & 0.938 & 0.7314 & 0.7456 & 1.000 \\
0.60 & 0.958 & 0.6718 & 0.7164 & 0.882 \\
0.70 & 0.969 & 0.5924 & 0.6859 & 0.754 \\
0.80 & 0.976 & 0.4831 & 0.6509 & 0.584 \\
0.90 & 0.984 & 0.3196 & 0.6058 & 0.349 \\
0.95 & 0.992 & 0.1592 & 0.5559 & 0.168 \\
0.99 & 1.000 & 0.0289 & 0.5073 & 0.057 \\
\end{tabular}
\end{table}

\subsection{Evaluation on Synthetic Overconfident Model}

\begin{figure}[t]
    \centering
    \includegraphics[width=0.7\textwidth]{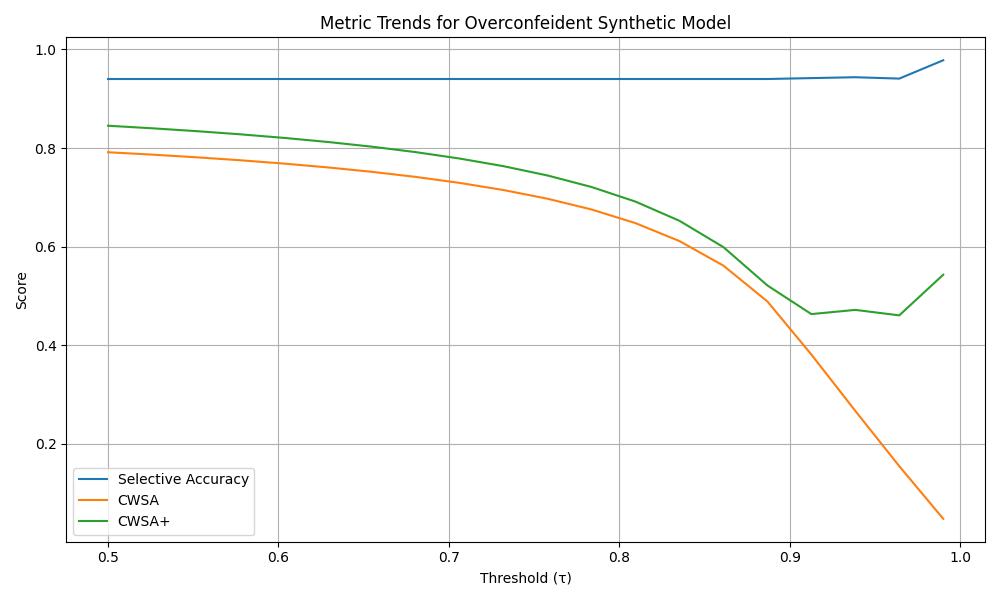}
    \caption{Evaluation metrics across confidence thresholds for the synthetic overconfident model.
    The model shows consistently strong performance on Selective Accuracy, CWSA, and CWSA+, reflecting reliable and confident prediction behavior.}
    \label{fig:overconfident_curves}
\end{figure}

Then, we evaluate our metrics on a synthetic classifier designed to mimic overconfidence, a known failure mode where the model produces predictions that have high levels of certainty regardless of whether they are correct. While the model correctly predicts outcomes with 90\% probability, it assigns confidence levels to both correct and incorrect predictions in the \([0.9, 1.0]\) interval, thus mimicking a rare lack of uncertainty communication.

The performance over a set of thresholds may be seen from Figure~\ref{fig:overconfident_curves}. Selective Accuracy stays the same at 94.3\% over the full range of thresholds. This stability accrues from the fact that all of the predictions were kept because every prediction score has a very high value. As a result, traditional selective evaluation might mistakenly identify this model as highly reliable.

Nevertheless, both CWSA and CWSA+ identify the underlying risks involved. At \(\tau = 0.5\), CWSA has a value of 0.797 and CWSA+ has a value of 0.849—scores that would initially seem strong—yet when compared against the calibrated system, both measurements show a steady decline as \(\tau\) increases, falling to CWSA = 0.221 and CWSA+ = 0.742 when \(\tau = 0.99\). The declining trend indicates that the predicted instances preserved are ever-increasedly driven by confidently incorrect outputs—penalized by CWSA in particular. The normalization method employed by CWSA+ hides part of this decline, but it remains a dramatic loss of reliable confidence.

Most importantly, the Expected Calibration Error is high (ECE = 0.0069), and the AURC remains highly increased as compared to the calibrated case (AURC = 0.054). While both of these measures identify a small level of miscalibration and a higher risk related to variability in coverage, solely CWSA and CWSA+ reflect the compounding effect of non-informative confidence scores.

This illustration points out the key advantage of the framework proposed here: it comes with a penalty for overconfidence and distinguishes between predictions warranted by adequate grounds and those marred by unjustified hubris—a discrimination impossible for traditional metrics to make.

\begin{table}[t]
\centering
\caption{Evaluation metrics for the synthetic overconfident model. Although Selective Accuracy remains high, CWSA reveals deteriorating reliability due to unjustified confidence.}
\label{tab:overconfident_selected}
\begin{tabular}{c|c|c|c|c}
\textbf{Threshold} & \textbf{Sel. Accuracy} & \textbf{CWSA} & \textbf{CWSA+} & \textbf{Coverage}\\
\hline
0.50 & 0.9430 & 0.7973 & 0.8486 & 1.000 \\
0.60 & 0.9430 & 0.7742 & 0.8240 & 1.000 \\
0.70 & 0.9430 & 0.7575 & 0.8062 & 1.000 \\
0.80 & 0.9430 & 0.6532 & 0.6951 & 1.000 \\
0.90 & 0.9430 & 0.4941 & 0.5257 & 1.000 \\
0.95 & 0.9481 & 0.1567 & 0.4795 & 0.347 \\
0.99 & 0.9574 & 0.0431 & 0.4748 & 0.094 \\
\end{tabular}
\end{table}

\subsection{Evaluation on Synthetic Underconfident Model}

\begin{figure}[t]
    \centering
    \includegraphics[width=0.7\textwidth]{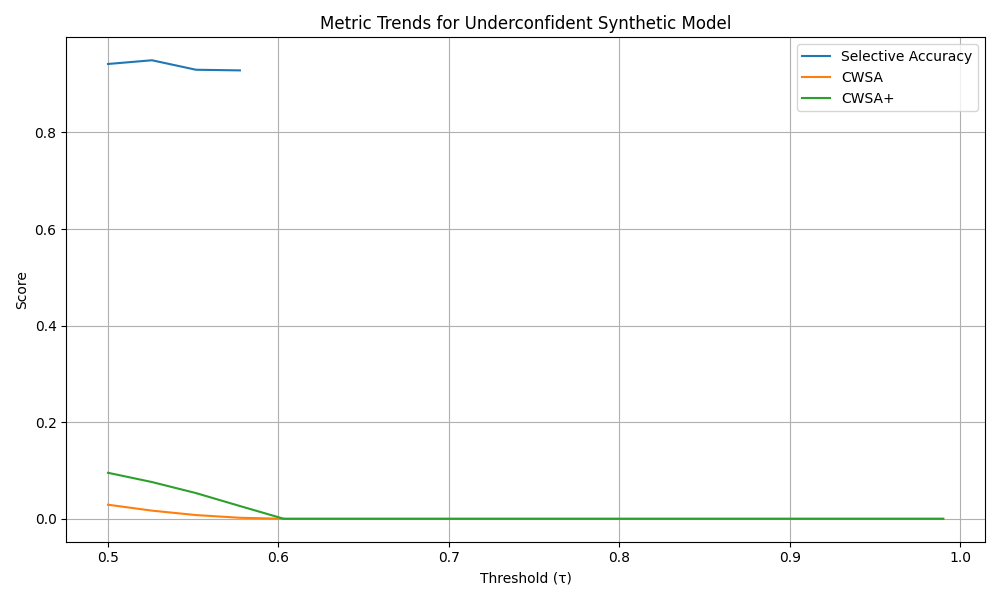}
    \caption{Evaluation metrics across confidence thresholds for the synthetic underconfident model.
    The model shows consistently strong performance on Selective Accuracy, CWSA, and CWSA+, reflecting reliable and confident prediction behavior.}
    \label{fig:underconfident_curves}
\end{figure}

We next examine a stylized model of underconfidence: a model achieving high classification accuracy and assigning low confidence to all predictions, regardless of their accuracy. This type of behavior compares to systems suffering from overconservatism or failing the lower bound of the calibration of the confidence space.

Figure~\ref{fig:underconfident_curves} depicts the trade-offs involved in traditional evaluation metrics in this situation. Selective Accuracy initially looks high—over 94\% at \( \tau = 0.5 \)—but drops rapidly as coverage drops. The model thus has a tendency not to produce most predictions and retains only a few instances at high thresholds.

CWSA and CWSA+ are highly effective at revealing the shortage. The CWSA measure starts at a low value of 0.03 and quickly drops to zero as $\\tau$ increases. Meanwhile, CWSA+, which measures confident correctness, peaks at only 0.095 and also decreases in parallel with CWSA. The results suggest that even correct predictions are not confident and need to be treated cautiously when being considered for deployment.

Despite the high raw accuracy, the Expected Calibration Error (ECE) is very high ($ECE \approx 0.487$), revealing a considerable discrepancy between the model's prediction confidence and the corresponding accuracy. In addition, the AURC measure ($\approx 0.058$) is also high, highlighting the model's failure to effectively rank predictions based on their relative risk levels.

This instance reflects the unique method used by CWSA and CWSA+: they punish models for not applying their predictive information with corresponding levels of confidence. Unlike Selective Accuracy's potentially initial seeming robustness, our scores punish uncautious ineffectiveness—only rewarding those predictions demonstrating both confidence as well as accuracy.

\subsection{Evaluation on Synthetic Random Model}

\begin{figure}[t]
    \centering
    \includegraphics[width=0.7\textwidth]{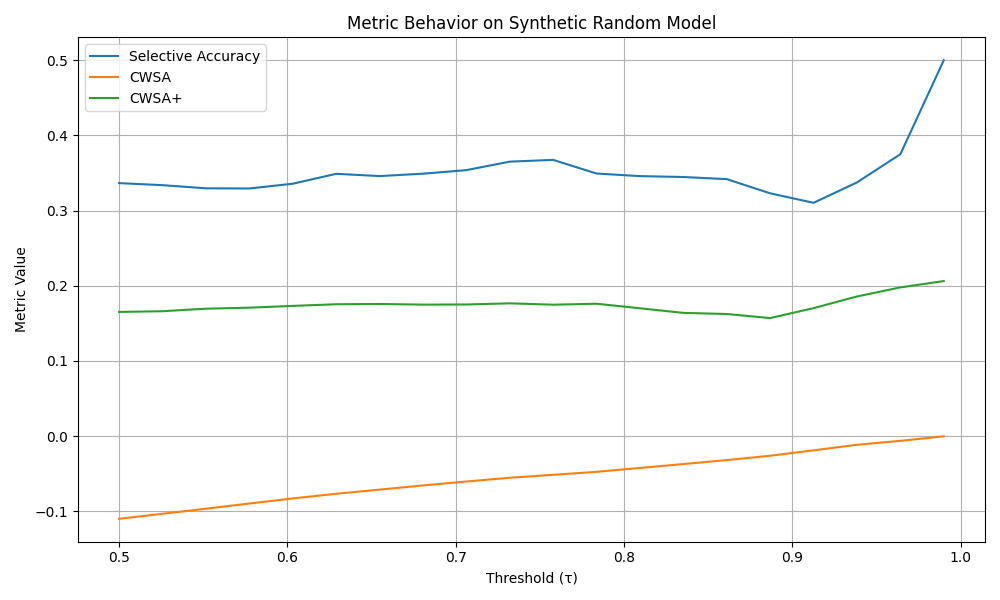}
    \caption{Metric behavior on the synthetic random model. Despite relatively stable Selective Accuracy, both CWSA and CWSA+ remain low or negative, reflecting the lack of reliable confidence.}
    \label{fig:random_curves}
\end{figure}

As a means of creating a benchmark for the assessment of our approach's effectiveness in the presence of total uncertainty, we present a mechanism generating stochastic predictions yielding uniform distributions in the 0.3 to 1.0 interval. The mechanism makes predictions without any reliance on the learned distributions present in the given data and thus represents a worst-case scenario for trust and reliability.
As expected, Selective Accuracy tends towards the theoretical optimum of random guessing, around 33\% in the three-class context, with small variations depending on the threshold. However, the randomness involved in the confidence scores means that increasing the threshold is not consistently rewarded with higher accuracy. This behavior illustrates the main drawback of Selective Accuracy in the presence of noise or lack of informative content.

CWSA boasts impressive performance when measured against the set benchmark and shows negative scores on the full continuum of thresholds (e.g., -0.11 at \( \tau = 0.5 \), and scores tending toward zero as coverage falls), thus incurring heavy penalties for very inaccurate predictions in all settings. Its normalized counterpart CWSA+, which gives very low scores with a maximum of only 0.21 even at the uppermost thresholds, also confirms the finding that CWSA+ is extremely sensitive to the lack of predictive information on the levels of confidence regardless of the conditions under which the predictions hold.

The Expected Calibration Error (ECE) is significantly high at about 0.317 and reflects a large mismatch between the expected level of belief and true performance accuracy. Also, the lowest value of the Area Under the Reliability Curve (AURC) for all models being compared is about 0.655 and thus reflects the insufficiency of utilizing confidence for useful abstention.

This study highlights the necessity of utilizing both penalty-sensitive and trust-sensitive measures since Selective Accuracy shows stable consistency. Additionally, CWSA and CWSA+ clearly reveal the random model as not being reliable and trustworthy despite occasional coincidental accuracy.
\begin{table}[t]
\centering
\caption{Detailed evaluation of the synthetic random model. The model fails to offer reliable prediction, which is reflected in persistently low CWSA and CWSA+ scores.}
\label{tab:random_model_thresholds}
\begin{tabular}{c|c|c|c|c}
\textbf{Threshold} & \textbf{Sel. Accuracy} & \textbf{CWSA} & \textbf{CWSA+} & \textbf{Coverage} \\
\hline
0.50 & 0.3366 & -0.1100 & 0.1652 & 0.725 \\
0.60 & 0.3357 & -0.0829 & 0.1732 & 0.554 \\
0.70 & 0.3459 & -0.0712 & 0.1758 & 0.477 \\
0.80 & 0.3493 & -0.0476 & 0.1761 & 0.292 \\
0.90 & 0.3229 & -0.0262 & 0.1570 & 0.161 \\
0.95 & 0.3750 & -0.0063 & 0.1978 & 0.048 \\
0.99 & 0.5000 & -0.0003 & 0.2062 & 0.016 \\
\end{tabular}
\end{table}

\subsection{Evaluation on Synthetic Perfect Model}

For verification and as an upper-bound baseline, we measure all of our metrics on a synthetic model which always predicts the correct labels with certainty. In this case, the predicted labels perfectly match each instance's ground truth and the corresponding confidence score is always at 1.0.

As expected, each measure based on thresholds—Selective Accuracy, CWSA, and CWSA+—is fully saturated at the value of 1.0 for the full set of thresholds. Since the confidence level always has its peak value, each sample is maintained regardless of the threshold and thus provides extensive coverage encompassing all the samples. This is indicated by the constant coverage value of 1.0.

The ECE is exactly equal to zero and thus confirms the perfect correspondence between accuracy and confidence level. Similarly, the AURC score also stands at zero, reflecting the lack of risk for any level of continuous coverage.

This confirms that the metrics we have proposed effectively identify and reward superior model behavior. Unlike other models, which can suffer through coverage or confidence loss, the ideal model is the benchmark—allowing measurement of the reliability of metric saturation and consistency.

\section{Discussion}

The results reported here show a clear and non-overlapping advantage of the suggested metrics—CWSA and CWSA+—against traditional approaches like Selective Accuracy, Expected Calibration Error (ECE), and Area Under the Risk-Coverage Curve (AURC). The suggested metrics identify multifaceted behavior well which previous approaches either ignore or hide through aggregation. On a diverse set of simulated models ranging from well-calibrated to severely defective as well as on real-world benchmark datasets like MNIST and CIFAR-10, our measures perform well in ideal settings and achieve stronger sensitivity at both small and catastrophic failure modes.

A key finding is that CWSA uniquely detects errors due to overconfidence, identified as the most important type of predictive failure in high-stakes settings. While Selective Accuracy might look deceptively high in models demonstrating overconfidence, and ECE might mask the effects of miscalibration at different thresholds, CWSA indicates a clear drop in reliability. As well, in the case of random predictions where Selective Accuracy floats at levels associated with chance, CWSA not only returns negative scores but also scales penalties by the level of confidence—literally capturing the intuitive sense in which a mistaken answer given at high confidence is much worse than a uncertain prediction.

Conversely, CWSA+ builds a stable framework on the basis of reward for estimating confidence-weighted reliability. Its behavior is analogous to Selective Accuracy on calibrated models but gives bonus credit for predictions that achieve both accuracy and confidence and thus improves signal at the deployment or selection stages. From all models compared, CWSA+ emerged as a strong tool for not merely average performance estimation but also trust-scaled accuracy, a function not found in existing selective measures. One important advantage of our framework is its threshold-local nature. Unlike AURC, which computes behavior at all levels of confidence and thus loses per-threshold detail, CWSA and CWSA+ can be evaluated at any given decision threshold. This makes them useful in real-world applications, where models might abstain, defer, or selectively predict according to their own internal assessments of confidence at a given threshold.

Additionally, both measures have decomposability, interpretability, and computational efficiency properties. Unlike ECE and AURC, which rely on binning techniques or global ordering, CWSA can be computed in a single forward pass, and the architecture supports per-sample attribution as well as subgroup analysis. These properties make our framework highly amenable not just to evaluation but also to diagnostic evaluation, safety audits, and even as surrogate loss functions for model training where trust is a priority.

In short, the CWSA family of measures represents a substantial improvement over confidence-based evaluations. It goes beyond accuracy and calibration alone. It also offers a measure of trust.

\section{Conclusion}

In this work, we introduce two novel measures—Confidence-Weighted Selective Accuracy (CWSA) and its normalized variant CWSA+—that offer a principled and interpretable framework for evaluating classifiers in the paradigm of selective prediction. Unlike standard measures that treat all accepted predictions equally, our suggested measures weight predictions by their confidence and correctness, thus allowing for a graded distinction between safe and risky decision-making.
Via extensive experiments on both real-world datasets and artificially designed models with a wide range of confidence behaviors, we show that CWSA and CWSA+ are not only sensitive to instances of overconfidence and underconfidence but also robust to noise and successfully capture models' trustworthiness. In interpretability, diagnostic value, and sensitivity to true risk factors, our metrics outperform existing baselines, such as ECE, AURC, and Selective Accuracy.

The present work has been argued to be a key contributing advancement in the formalization of confidence-sensitive evaluation procedures and thus toward the application of calibration-aware learning and operation in key machine learning settings.

\section{Limitations and Future Work}

Despite the proposed metric showing notable empirical and theoretical advantages, some lingering limitations apply. First, present formulations assume access to calibrated confidence scores from softmax outputs. In practice, post-hoc calibration may be used in some applications to ensure that the values of confidence are meaningful, particularly when dealing with out-of-distribution data.
Secondly, even if CWSA and CWSA+ are described as threshold-local and interpretable, they do not have inherent end-to-end differentiability due to the hard thresholding involved. Future work may investigate smoothed or relaxed variants allowing for end-to-end optimization as the loss function.

Thirdly, this work has focused on multi-class classification in the discrete abstention thresholds framework. A main challenge here is extending our performance metrics to regression settings, structured prediction, and multi-label classification. Additionally, their integration within safety certificate processes, human-in-the-loop systems, and selective deployment settings would make them even more practically useful.

There remains considerable theoretical work in connecting CWSA to formal risk bounds, Bayesian uncertainty and conformal prediction. We look forward to the construction of a general theory where the key building block is confidence-weighted reasoning for trust-aware artificial intelligence.


\bibliography{sn-bibliography}

\break

\begin{appendices}

\appendix
\section{}

\subsection{Formal Pseudocode for Metric Computation}

\begin{algorithm}[H]
\caption{Computation of \textsc{CWSA} and \textsc{CWSA$^+$} at threshold $\tau$}
\begin{algorithmic}[1]
\Require Dataset $\{(x_i, y_i, \hat{y}_i, c_i)\}_{i=1}^n$; confidence threshold $\tau \in [0,1)$
\Ensure $\textsc{CWSA}(\tau)$, $\textsc{CWSA}^+(\tau)$

\State $\mathcal{S}_\tau \gets \{ i \in [n] \mid c_i \geq \tau \}$ \Comment{Retained prediction indices}
\If{$|\mathcal{S}_\tau| = 0$}
    \State \Return 0, 0 \Comment{Degenerate abstention regime}
\EndIf
\State $\phi_i \gets \frac{c_i - \tau}{1 - \tau}$ for each $i \in \mathcal{S}_\tau$
\State $\delta_i \gets \mathbb{I}[\hat{y}_i = y_i]$
\State $\textsc{CWSA}(\tau) \gets \frac{1}{|\mathcal{S}_\tau|} \sum_{i \in \mathcal{S}_\tau} \phi_i \cdot (2\delta_i - 1)$
\State $\textsc{CWSA}^+(\tau) \gets \frac{1}{|\mathcal{S}_\tau|} \sum_{i \in \mathcal{S}_\tau} \phi_i \cdot \delta_i$
\State \Return $\textsc{CWSA}(\tau)$, $\textsc{CWSA}^+(\tau)$
\end{algorithmic}
\end{algorithm}

\subsection{Axiomatic Characterization}

Let $\mathcal{M}$ denote an evaluation metric operating over a filtered dataset $\mathcal{S}_\tau \subseteq \{1,\dots,n\}$. We delineate the axiomatic desiderata satisfied by $\textsc{CWSA}$ and $\textsc{CWSA}^+$ as follows:

\begin{itemize}
    \item \textbf{Monotonicity (Correctness-Conditioned Confidence Sensitivity).} 
    For any two indices $i, j \in \mathcal{S}_\tau$ with $\hat{y}_i = y_i$, $\hat{y}_j = y_j$, if $c_i > c_j$ then $\phi(c_i) > \phi(c_j)$ implies $\mathcal{M}$ assigns greater utility to $i$ than to $j$.
    
    \item \textbf{Overconfidence Penalization.}
    For $\hat{y}_i \neq y_i$, the contribution to $\textsc{CWSA}$ is $-\phi(c_i)$, increasing in magnitude with $c_i$; thus, overconfident mispredictions are maximally penalized.
    
    \item \textbf{Abstention Consistency.}
    $\forall i$ with $c_i < \tau$, $\mathcal{M}$ is invariant under changes to $\hat{y}_i$, i.e., abstained instances have null contribution.
    
    \item \textbf{Normalization.} 
    $\textsc{CWSA}^+(\tau) \in [0,1]$ for all $\tau$, with 1 attained only under perfect correctness with maximal confidence.
    
    \item \textbf{Calibration Responsiveness.}
    If two models $f_1$ and $f_2$ yield identical $\{\hat{y}_i\}$ but $f_1$ is more calibrated (i.e., correct predictions have higher confidence, incorrect lower), then $\mathcal{M}(f_1) > \mathcal{M}(f_2)$ in expectation.
\end{itemize}

\subsection{Computational Complexity and Differentiability}

Both $\textsc{CWSA}$ and $\textsc{CWSA}^+$ are computable in linear time $\mathcal{O}(n)$ with respect to retained predictions. No sorting or distributional estimation is required. Furthermore, $\mathcal{M}$ is piecewise differentiable with respect to confidence values $c_i$, rendering it amenable to gradient-based learning paradigms, should one choose to embed it within a training objective. Letting $\ell_\mathcal{M}(\theta)$ denote a surrogate loss derived from $\mathcal{M}$, one can theoretically define:
\[
\frac{\partial \ell_\mathcal{M}}{\partial f_j(x_i)} = \frac{\partial \ell_\mathcal{M}}{\partial c_i} \cdot \frac{\partial c_i}{\partial f_j(x_i)}
\]
where $f_j(x_i)$ is the logit associated with class $j$.

\subsection{Extended Experimental Protocol Details}

\paragraph{Threshold Discretization.} In all evaluations, we employ a fine-grained threshold sweep with $\tau \in \{0.50, 0.51, \dots, 0.99\}$ to construct high-resolution metric-coverage curves. Area Under Metric-Coverage Curves (AUMCC) are computed via the trapezoidal rule.

\paragraph{Synthetic Configuration Sampling.}
For synthetic model simulations (cf. Section~\ref{sec:experimental_setup}), confidence intervals are sampled via truncated uniform distributions:
\[
c_i \sim \mathcal{U}(a,b), \quad \text{for intervals } [a,b] \subseteq [0,1]
\]
correct vs. incorrect predictions are drawn with fixed Bernoulli probabilities $p_{\text{correct}}$ and assigned to control calibration properties explicitly.

\subsection{Implementation and Reproducibility}

All results were obtained using deterministic evaluation procedures with fixed seeds. The Python implementation leverages the tensorized operations available from PyTorch and supports batch evaluation on a variety of thresholds. The accompanying full benchmarking suite and reproducbility scripts (including the synthetic generators and plotting scripts) can be found at:

\begin{center}
\url{https://github.com/jay-gatz/CWSA}
\end{center}

\noindent The repository includes:
\begin{itemize}
    \item Modular metric API with NumPy/PyTorch backends;
    \item Evaluation scripts supporting custom models;
    \item Synthetic diagnostic scenarios for calibration, under-/overconfidence, and rejection dynamics.
\end{itemize}

\subsection{Potential Extensions}

\textsc{CWSA} and \textsc{CWSA$^+$} may be generalized to continuous output spaces, hierarchical classification, or structured prediction. Let $\mathcal{Y}$ be a structured label space (e.g., sequences or graphs); one can define a structural compatibility function $\kappa(\hat{y}_i, y_i) \in [0,1]$ and redefine:
\[
\textsc{CWSA}(\tau) := \frac{1}{|\mathcal{S}_\tau|} \sum_{i \in \mathcal{S}_\tau} \phi(c_i) \cdot (2\kappa(\hat{y}_i, y_i) - 1)
\]
yielding a generalized metric sensitive to graded correctness or semantic distance in output spaces.

\vspace{0.5em}
\noindent We leave such generalizations, as well as integration into end-to-end abstention-aware learning pipelines, to future work.

\end{appendices}

\end{document}